\documentclass[conference,two column]{IEEEtran}
\IEEEoverridecommandlockouts
\usepackage{cite}
\usepackage{amsmath,amssymb,amsfonts}
\usepackage{algorithmic}
\usepackage{graphicx}
\usepackage{textcomp}
\usepackage{xcolor}
\def\BibTeX{{\rm B\kern-.05em{\sc i\kern-.025em b}\kern-.08em
    T\kern-.1667em\lower.7ex\hbox{E}\kern-.125emX}}

\newcommand{\linebreakand}{%
      \end{@IEEEauthorhalign}
      \hfill\mbox{}\par
      \mbox{}\hfill\begin{@IEEEauthorhalign}
    }

\begin{document}

\bibliographystyle{IEEEtran}

\title{A New Sentence Ordering Method Using BERT Pretrained Model} 

\author{
\IEEEauthorblockN{Melika Golestani}
\IEEEauthorblockA{\textit{department of ECE} \\
\textit{University of Tehran}\\
Tehran, Iran \\
melika.golestani@gmail.com}
\and
\IEEEauthorblockN{Seyedeh Zahra Razavi}
\IEEEauthorblockA{\textit{department of Computer Science} \\
\textit{University of Rochester}\\
Rochester, USA \\
srazavi@cs.rochester.edu}
\and
\IEEEauthorblockN{Heshaam Faili}
\IEEEauthorblockA{\textit{department of ECE} \\
\textit{University of Tehran}\\
Tehran, Iran \\
hfaili@ut.ac.ir}

}

\maketitle

\begin{abstract}
Building systems with capability of natural language understanding (NLU) has been one of the oldest areas of AI. An essential component of NLU is to detect logical succession of events contained in a text. The task of sentence ordering is proposed to learn succession of events with applications in AI tasks. The performance of previous works employing statistical methods is poor, while the neural networks-based approaches are in serious need of large corpora for model learning.
In this paper, we propose a method for sentence ordering which does not need a training phase and consequently a large corpus for learning. To this end, we generate sentence embedding using BERT pre-trained model and measure sentence similarity using cosine similarity score. We suggest this score as an indicator of sequential events’ level of coherence. We finally sort the sentences through brute-force search to maximize overall similarities of the sequenced sentences. Our proposed method outperformed other baselines on ROCStories, a corpus of 5-sentence human-made stories.
The method is specifically more efficient than neural network-based methods when no huge corpus is available. Among other advantages of this method are its interpretability and needlessness to linguistic knowledge.
\end{abstract}

 %کیوردهای خوبی نزدم، اگه ممکنه ادیتشون کن low resource ordering خوبه مثلا؟.

\section{Introduction}
Modeling the construct of coherent texts and logical sequence of events is an underlying issue in processing natural language. By "events" in natural languages, we particularly mean verbs of sentences and their associates such as objects and subjects, since verbs indicate events occurring in sentences \cite{chambers2009unsupervised}. Modeling event coherence and sequence in a text is intuitively associated with logical consistency and topic transition of the sentences, as in a good text, the events occurring in each sentence command a logical construct and a particular subject. 
%Zahra: cant understand this sentence %i changed it
Sentence ordering \cite{barzilay2008modeling} is a proposed subtask  of  coherence  modeling.
“Sentence ordering” means that we retrieve the original paragraph or story from the text shuffled sentences. This includes learning some coherence between the sentences, regardless of their input shuffled sequence. So, the model needs to identify crucial properties that cause text coherence. This is a notable task in machine intelligence, since incorrect sequence of sentences generated by an AI severely declines the user’s comprehension. It also gives ways to computers to better understand longer natural language texts such as stories, reports and narratives.

Our main objective is to formulate models that could capture the sequence of events through ordering and sequencing a set of sentences. More precisely, we seek the correct sequence of disordered sentences of a story. Our specific task in this paper is to sequence the disordered sentences of short stories and discover the original correct ordering. The stories come from the ROCStories dataset (Mostafazadeh, Chambers, et al., 2016), a corpus of human-made 5-sentence stories. The intuition we get from this work can help AI systems better understand and predict the nature of connections between the events. The following are among applications of this task:
\begin{itemize}
    \item In text generation systems, the text would be generated more coherently \cite{lapata2003probabilistic}.
    \item It can prevent system failure due to an increase in the number and length of turns in dialogue systems \cite{roemmele2018neural}.
    \item It is highly effective in systems such as extractive text summarization \cite{barzilay2002inferring}, story-telling text completion recommender \cite{roemmele2015creative}, retrieval-based question answering (QA) \cite{yu2018qanet} where the incorrect sequence of sentences may lead to a weak performance of the system and decline users' understanding.
\end{itemize}
The previous works fall into two traditional and deep learning approach categories. Traditional methods including rule-based and statistical types are of low quality, where the latter requires a large corpus for model learning. We propose a method that does not need any training dataset as it needs no training phase, but instead relies on the language and semantic features.

Our method has three components:
\begin{itemize}
    \item Sentence embedding with pretrained SBERT-WK \cite{wang2020sbert},
    \item Computing scores based on cosine similarities between each pair of sentence embeddings \cite{sravanthi2017semantic},
    \item Sentence ordering using brute-force search.
\end{itemize}

The essential initiative of this study is that it recommends an interpretable method needless of training corpus due to the absence of training phase in carrying out this task.
We implemented the proposed method on a corpus of short human-made commonsense stories, called ROCStories \cite{mostafazadeh2016corpus}. We showed that the proposed sentence similarity method can outperform the previous statistical methods, while it does not need any training data in contrast to neural networks models. 

The remainder of this paper is structured as follows. Section II looks through relevant previous work in the area. In section III we represent a formal description of the task. We introduce the dataset and explain our method in section IV and section V respectively. The baselines are described in section VI, then the results are discussed in section VII. We wrap up the paper with a conclusion in section VIII.

\section{Related work}
Previous works on the sentence ordering task could be divided into two categories: Traditional approaches and Deep learning approaches.

\subsection{Traditional Approaches}
This group consists of rule-based and probabilistic methods. Previous work on coherence modeling mainly focused on the utilization of linguistic features and statistical models. Rule-based methods are obsolete for many reasons, such as lack of portability, lack of scalability, the urgent need for language knowledge, and the high cost of generating rules. As a result, probabilistic methods were presented. For instance in \cite{lapata2003probabilistic} transition probabilities between sentences are calculated and the sentence orders are decoded greedily. Topic and topic transition modeling with Hidden Markov Model (HMM) is used in \cite{barzilay2004catching}. In another approach presented in \cite{barzilay2008modeling}, they extract entities and learns entity transition probabilities.
Probabilistic methods have some degree of portability and scalability, but still need language knowledge and do not have a remarkable quality. Yet, the biggest limitation of probabilistic methods is their reliance on linguistic domain knowledge. So, deep learning approaches were suggested.

\subsection{Deep Learning Approaches}
%این سه تا موردی که گفتم رو میشه جمله بندیشو عوض کنی لطفاا
With the increase in computing power, deep learning methods have found their place and have been used to solve many problems of Artificial Intelligence. Recently, neural network-based models have shown powerful capability in sentence ordering. For example, \cite{gong2016end}
investigated the effectiveness of various neural models on discovering the correct order of each sentence pair. Multiple neural network models based on individual and pairwise element-based predictions (and their ensemble) have been implemented in \cite{agrawal2016sort}. \cite{li2016neural} applied sequence-to-sequence based generative models, introduced in \cite{sutskever2014sequence}, to model pairwise coherence. In \cite{chen2016neural} used neural networks and pairwise ordering for sequencing sentences are used. %اینا هم باید جمله بندیش عوض شه یا خودم کلا توضیح بدم. از جایی برش داشتم الان.
During the past few years, story understanding has been attracting attention in NLP research. More specifically, using neural networks has been of interest to solve the sentence ordering task (for example: \cite{prabhumoye2020topological}, \cite{kumar2020deep}, \cite{yin2020enhancing}, \cite{yin2019graph}, \cite{wang2019hierarchical}, \cite{oh2019topic}. 
%and \cite{logeswaran2016sentence}), 
In addition, in \cite{bohn2018learning} sentence embeddings have been created that include text coherence models.

Although deep learning-based methods have shown success in tackling problems such as sentence ordering, this should not disadvantage the previous methods. One issue with neural network models is their need to  enormous amount of training data. These methods show very poor performance when data is limited or they overfit and severely lose their quality. The other disadvantage of these methods is their time complexity. In addition, deep learning methods have very little interpretability and it is hard to understand on what basis and with what features they solve the problem. These issues persuade researchers to not easily leave statistical approaches. In this work, we show that our method can outperform neural network based models where limited data is available.

\section{Task description}
Suppose S is the set of $n$ unordered sentences came from a coherent text, namely:

\begin{equation}
S = {s_1}, {s_2}, \dots, {s_n}  
\end{equation}

A sentence ordering, $o$, would be a permutation of sentences 1 to n:

\begin{equation}
s_{o_1}> s_{o_2} > \dots > s_{o_n}
\end{equation}

The task of sentence ordering aims to find the correct permutation of sentences, it means a permutation $o^*$ that fits the original ordering of the text in gold data: 

\begin{equation}
s_{o_1^*}> s_{o_2^*} > \dots > s_{o_n^*}
\end{equation}

Based on the above definition and notions we propose our sentence ordering method.

\section{Dataset}

\subsection{ROCStories}

The data we use in this study comes from a corpus of commonsense stories. The corpus called ROCStories consists of 98,162 stories, where each story has exactly five sentences. The average number of words per story is 50. Among all, 3,742 of the stories have two choices for the final sentence, where only one of them makes sense as the fifth sentence. The data made by human and designed for the LSDSem'17 shared task  \cite{mostafazadeh2017lsdsem}. The task is to propose methods that can decide the correct final sentence of a story, having the first four \cite{mostafazadeh2016corpus}. 

Important features of this dataset that make it useful in tasks related to learning events sequences and ordering sentences are:

\begin{itemize}
    \item The stories are full of stereotypical causal and temporal relations between events, making them a great resource for commonsense reasoning and generic language understanding. Besides, this makes it possible to learn the narrative structure in a wide range of events.

    \item The dataset contains a high-quality collection of realistic and non-fictional short stories that can be used to train coherent text generation models.

    \item The stories do not include anything irrelevant to the core story.
    
    \item Since publishing, the dataset has been used in not only LSDSem'17 shared task, but also in other tasks relevant to story understanding and natural language processing.
\end{itemize}

These features make the corpus a valuable one for analyzing sentence ordering in a text. As mentioned, the results can be used in solving other tasks that need to generate coherent texts and improve their output.

\section{Recommended Method Called Sentence Correlation Measure (SCM)}

We recommend a sentence ordering method that relies on the sentences' semantic similarities. To that end, the similarity between sentences should be measurable. So, we use vector representation of language which is very useful in placing words and sentences in a measurable vector space. 

\begin{figure}[h!]
  \includegraphics[scale=0.40]{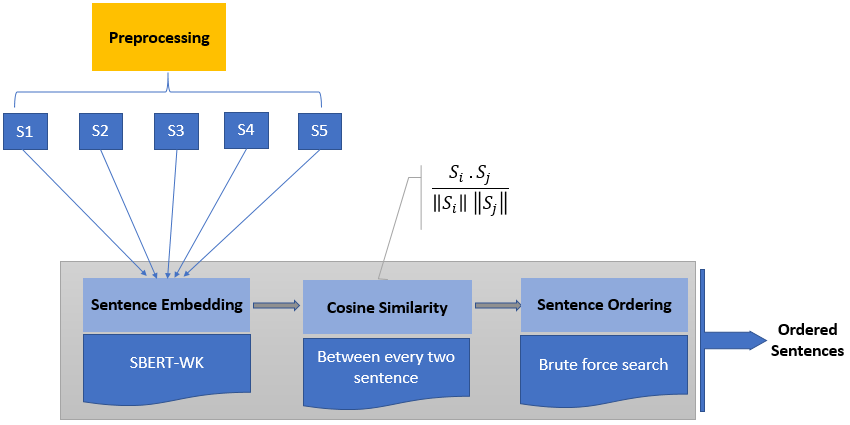}
  \caption{Recommended method: Sentences' Correlation Measure (SCM). In the first component, the sentences are encoded to vectors. We used pre-trained SBERT-WK model for that. Then cosine similarity is used as the scoring function. At the end, sentences are ordered by brute-force search. }
  \label{figure 1}
\end{figure}

As shown in the Figure \ref{figure 1} Our method has three components: sentence encoder, scoring function, and sentence organizer. For each 5-sentences story from the dataset, we fist encode the sentences using a pre-trained model of SBERT-WK \cite{wang2020sbert}. The encoding has a preprocessing step, where pronouns are replaced by their associated entities using the neural coreference system \cite{clark2016improving}. After embedding the shuffled input sentences, we measure the similarity between each pair of sentences by calculating the cosine similarity \cite{sravanthi2017semantic} between them. In other words, we measure the similarity between each sentence and the other four sentences of the shuffled set of each story to calculate the semantic similarity of two sentences. In \cite{zhelezniak2019correlation} it is shown that using cosine similarity to measure the similarity and correlation of two sentences is a good method.

To sequence the sentences, we perform the brute-force search methods with aiming to maximize the total similarity of consecutive sentences, as follows:

\begin{equation}
{\sum_{i=1,...,4}}{S({{s_i}_O},{{{s_i}_+1}_O})} = {\sum_{i,j=1,...,5} \&  i\neq j}{S({s_i},{s_j)}}
\end{equation}

Where $S({s_i}, {s_j})$ is equal to the cosine similarity between $s_i$ and $s_j$ and ${{s_i}_O}$ is the i-th sentence in the output ordering. $s_i$ is the i-th sentence in the non-ordered permutation of  sentences.
Based on sentence similarity values, we develop six distinct methods for sentence ordering and report the results. We also tried Universal Sentence Encoder \cite{cer2018universal} as the sentence encoder. We also use pre-trained BERT word embeddings \cite{devlin2018bert} and following \cite{jiang2020combining}, we compute $r_{i,j}$, the similarity between two sentences i and j, as follows:

\begin{equation}
r_{i,j} = 
\frac{\sum_{w\epsilon s_i}{S(w,s_j)}}
    {2\left|s_i\right|}
    +
\frac{\sum_{w\epsilon s_j}{S(w,s_i)}}
    {2\left|s_j\right|}    
\end{equation} 
where $S(w, s_j)$ is the maximal cosine similarity between input word w and any words in the sentence $s_j$. In this case, we encode the words individually, so we do not map the sentences into vectors.

A nearest neighbor search the sentence organizer and reports the results. The nearest neighbor search is computed by:

\begin{equation}
Max(S(s_x, s_i  , i=1,2,3,4,5 , i\neq x)) = S(s_x , s_y)
\end{equation}

\begin{equation*}
{{Next-Sentence}_x}= s_y | (Max(S({{s_x}, {s_i}}
\end{equation*}
\begin{equation}
i=1,2,3,4,5 \& i\neq x)) =S(s_x , s_y))
\end{equation}

 % I added these lines here:
 % not clear what that is اینجا گفتم دو تا روش بوده که یکی همپوشانی nکرم هاست و اون یکی مپ کردن جملات به بردارهایی مه میانگین بردارهای کلمات این جمله هستن. این روش ها بیس لاین هستن ولی روی ترتیب دهی به جملات ترین و تست نشده بودن. من این کار رو کردم.
We also developed sentence n-gram overlap \cite{mostafazadeh2016corpus} and Continues Bag of Words (CBoW) for sentence ordering, both explained in the next section.

%شاید بهتر باشه این شکل رو تک ستونه قرار بدیم که بزرگتر باشه.

\section{Baseline}
 % I wrote this section again
To evaluate our method performance we use four baselines:
\begin{itemize}
\item using Continues Bag of Words (CBoW) for sentence embedding: CBoW  word embedding was proposed by \cite{mikolov2013efficient}. CBoW sentence embedding calculates the sentence vectors by taking the average of the embeddings of the words in a sentence. As a baseline, we replace our SBERT-WK encoder with CBoW and employ cosine similarity of all pairs of sentences.

\item Seq2Seq+Pairwise \cite{li2016neural}: This is a pairwise model which predicts the next sentence given the current sentence. Approaches like this, capture the local coherence of text based on the neural networks.

\item SkipThought+Pairwise \cite{agrawal2016sort}: This method involves combining two points: 1. the unary model of sentences without considering the contextual text, and 2. the model of the pair of sentences according to the context that encodes the relative composition of the sentences. In this paper, the skip thought vectors \cite{kiros2015skip} has been used to map sentences to vector space. This model takes as input a pair of SkipThought sentence embeddings and outputs the pairwise ordering of the input sentences.

\item sentence n-gram overlap: In \cite{mostafazadeh2016corpus} the use of n-grams overlaps has been suggested to select the fifth sentence as the final sentence of the five-sentence Rocstories stories from the proposed two man-made choices. Here, we use this method to arrange the sentences. In this method, the sentence that overlaps with the current sentence in the most n-grams will be selected as the next sentence of the current sentence. Overlap measurements of up to 4-grams were measured using Smoothed-BLEU \cite{lin2004automatic}.

Also, we compare our results with several commonly-used proposed models:

\item LSTM+PtrNet \cite{gong2016end}: This is an end-to-end approach that uses pointer network for ordering. It treats the out-of-order set of sentences as a sequential input for encoder and predicts sentence orders recurrently.

\item LSTM+Set2Seq \cite{logeswaran2016sentence}: This method uses an encoder-decoder architecture based on LSTM and proposed a pairwise model followed by a beam search decoder to seek the ground truth sentence order.

\item ATTOrderNet \cite{cui2018deep}: This model has three components: sentence encoder, paragraph encoder, and paragraph decoder. For sentence encoding it uses Bi-LSTM followed by attention layer, then for paragraph encoding it uses a feed forward self-attention mechanism. Compared to previous models, it is less sensitive to the permutation of input sentences. The general idea is to learn joint embedding of a group of sentences.

\item Hierarchical Attention Neural Network (HAN) \cite{wang2019hierarchical}: The goal is to capture word clues and dependencies between sentences. Its encoder and decoder are based on self-attention mechanism.

\item SE-Graph \cite{yin2019graph}: This model proposed to develop ATTOrderNet \cite{cui2018deep}. Comparing to ATTOrderNet, it adds entities' information. As the model does not have a feedforward architecture, non-consecutive sentences are not connected, so it conveys less noise.

\item Enhancing PtrNet + Pairwise  \cite{yin2020enhancing}: Its encoder is based on Bi-LSTM. It is a pairwise model which uses Pointer Network for sentence ordering.

\end{itemize}

We re-implement these methods on a part of ROCStories to demonstrate the effect of lack of data on model training.

\section{Evaluation Metrics}

To evaluate the proposed ordering, we need some metrics. We use two commonly used metrics: Kendall's tau and PMR, both explained below.

\subsection{Kendall's tau ($\tau$)}
\begin{figure} [h!]
  \centering
 
  \includegraphics[scale=0.35]{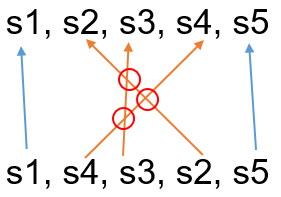}
  \caption{showing the inversion. $s_i$ is the ith sentence in the correct permutation of sentences and the out put permutation is $s_1$,$s_4$,$s_3$,$s_2$,$s_5$. The red circles indicate inversions. }
  \label{figure 2}
\end{figure}

Kendall’s $\tau$ measures the ordinal association between two sequences to automatically evaluate coherence modeling as follows:

\begin{equation}
\tau = 
1-
\frac{(2*number\_of\_inversions)}
    {N*(N-1)/2}    
\end{equation}

where N denotes the length of the sequence (i.e. the number of sentences of a story). The function inversions return the number of sentence-pair interchanges for reconstructing the correct order from the predicted order which is shown in the Figure \ref{figure 2}.
Kendall's $\tau$ value ranges from -1 to 1. A higher value indicates better performance. The evaluation metric closely correlates with users' ratings and reading times, which are related to readability. So, \cite{lapata2006automatic} suggests that Kendall's $\tau$ score for sentence ordering correlates with human judgement.

\subsection{Perfect Match Ratio (PMR)}
The perfect match ratio is the fraction of the number of exactly matching orders, so no partial credit is given for any incorrect permutations. Mathematically, it can be written as follows:

\begin{equation}
PMR = 
\frac{(number\_of\_correct\_pairs)}
    {N*(N-1)/2}    
\end{equation}

where N is the length of the sequence. PMR value ranges from 0 to 1 where 1 indicates that the proposed order matches the correct order, and the PMR value would be 0 if the predicted order is exactly the opposite of the correct order.

\section{Experimental Results}
As mentioned in section V, we develop six distinct methods based on sentence similarity for sentence ordering with variant components to compare and propose the best one. 
To this end, we use SBERT-WK and USE for sentence encoder (the first sentence embeddings has a dimension of 768 and the others are 512), cosine similarity for the scoring function, and brute-force search and nearest neighbor search for generating the ordering. Besides, we consider a word encoder via  BERT word embedding, then we compute the similarities between two sentences using equation (3). The results are shown in the Table 1. %بعدا رفرنس دادن به معادله رو درست میکنم

\begin{table}[htbp]
\caption{SCM results by changing each component. }
\begin{center}
\begin{tabular}{|c|c|c|}
\hline
\textbf{Method Components }&\textbf{Tau}&\textbf{PMR} \\
\hline
\textbf{SCM (SBERT-WK + Brute Force Search)} & \textit{0.5582}& \textit{0.2495} \\
\hline
\textbf{SBERT-WK + Nearest Neighbor Search} & \textit{0.5010}& \textit{0.1670} \\
\hline
\textbf{USE + Brute Force Search} & \textit{0.5833}& \textit{0.2300}  \\
\hline
\textbf{USE + Nearest Neighbor Search} & \textit{0.4879}& \textit{0.1474}  \\
\hline
\textbf{BERT word embedding + Brute Force Search} & \textit{0.5502}& \textit{0.2342}  \\
\hline
\textbf{BERT word embedding + Nearest Neighbor Search} & \textit{0.4817}& \textit{0.1696}  \\
\hline
\end{tabular}
\label{results}
\end{center}
\end{table}

It can be seen that the best result belongs to SCM method with spectral vectors of SBERT-WK sentences and brute-force search sorting. Based on the results, the best-case scenario of the first proposed method, which we call SCM (SBERT-WK + BFS), is: 

1. Replacing pronouns by entities using the neural coreference system \cite{clark2016improving}, 

2. Creating sentence embeddings using the pre-trained SBERT-WK model, 

3. Scoring all pairs of sentences by applying the cosine similarity scoring function to the sentence vectors, 

4. Determining the order by brute-force search method to maximize the sum of similarities of each consecutive sentence pair.

As can be seen in the results of this method, the brute-force search method always works better than the nearest neighbor search method. This happens for two reasons:

1. The nearest neighbor search is a greedy one. As a result it is possible for it to stop at the local optimum \cite{cormen2009introduction}.

2. In this case, the method is not able to identify which of the sentences is the initial sentence, and always considers the sentence that is in the first place in the input as the first sentence, but this will not be the case in the brute-force search.
In the next table, the results of the proposed method are compared with the results of the previous methods.

\begin{figure}[h!]
  \includegraphics[scale=0.7]{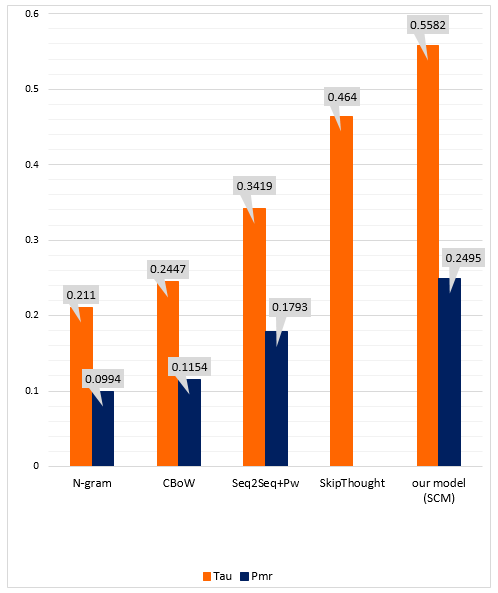}
  \caption{Comparing the SCM result versus the basic methods. SCM performs significantly better than the baselines in both criteria, while our method has no training. }
  \label{figure 3}
\end{figure}

Figure \ref{figure 3} compares the SCM result with the basic methods. The values reported for all methods on ROCStories dataset. We randomly split the dataset by 8:1:1 to get the training, validation and testing data of 78529, 9816 and 9817 stories respectively. The results show the usefulness of using vector space to display text. The $\tau$ criterion for the basic statistical method, which uses the N-gram subscription of sentences to sort them is 21.10\%. However, the same criterion for other basic methods based on the use of word spectral vectors and next sentence predictions is equal to 0.2447, 34.19\% and 46.40\%. We have been able to achieve the $\tau$ of 55.82\% by arranging the sentences using the similarity of the SBERT-WK sentence embeddings as brute-force search. This improvement is also true for the PMR criterion and We have been able to achieve the PMR of 24.95\%.

The results show that our proposed method is able to perform significantly better than the baselines in both criteria. The method does not include a training phase, in contrary to some of the baselines. This is the main strength of the proposed method. If no or limited training data is available, then SCM will be the most efficient method. Another advantage of this method is that it is interpretable, while neural networks-based methods usually are not. 

\begin{figure}[h!]
  \includegraphics[scale=0.7]{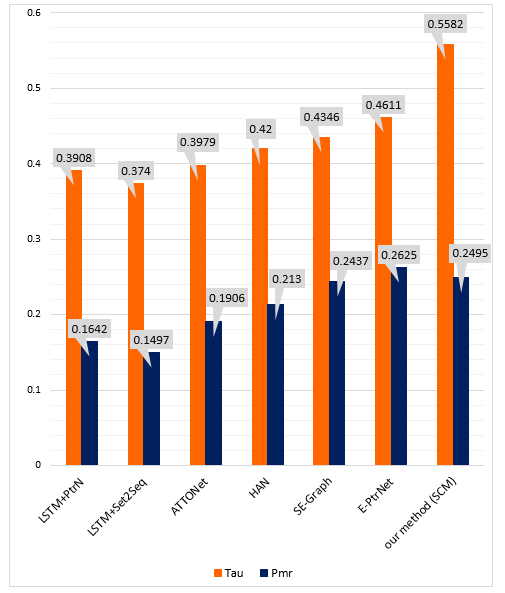}
  \caption{Comparing SCM with deep learning-based methods. These models are trained on a part of ROCStories to show the importance of access to a big corpus for using these methods which are based on deep learning. SCM doesn't any training, so SCM does not require a training dataset.}
  \label{figure 4}
\end{figure}

Figure \ref{figure 4} shows the results of SCM compared to neural networks-based methods. We have trained these methods on a portion of ROCStories to demonstrate the effect of the small size of data on performance of these methods. We make a random split on the dataset to get the training, validation and testing datasets of 19,632, 2,454 and 2,454 stories, respectively. We aim to demonstrate the power of our proposed method in the presence of small training dataset. It is clear that with no training dataset, deep learning methods and other machine learning methods can not be used. Our proposed method dramatically improves the methods that have no learning phase.

The results show that SCM wins the competition if there is no access to a large training dataset, whereas new methods with their complex architectures fail. We could have used less data to train the models to make them even less efficient, but our goal was to show that even a fair amount of data cannot solve the complex problem of sentence ordering, while at the same time, SCM shows reasonable performance using no training data.

\section{Conclusion}
In this paper, a method for sentence ordering called SCM was presented. The method does not have a training phase. Instead, SCM uses three components: 1- a sentence encoder based on the pretrained SBERT-WK model which maps sentences into a vector space to create a conceptual representation of them. 2- a scoring component that using the cosine similarity function, assigns points to each pair of sentences. 3- an ordering component that generates an ordering based on a brute-force search to maximize the total cosine similarities of every two consecutive sentences.

SCM significantly outperforms the baselines. Also due to the lack of training phase, this method does not require a training corpus, and therefore is the best choice in the absence of training dataset. Moreover, when there is a training corpus that is small for training deep learning models, the quality of such methods will be drastically reduced. So, SCM has a superiority. SCM has also a high interpretability comparing to neural network methods, which enables researchers to study the causes of its deficiencies and improve its performance.

\bibliography{paper2} 

\begin{thebibliography}{10}
\providecommand{\url}[1]{#1}
\csname url@samestyle\endcsname
\providecommand{\newblock}{\relax}
\providecommand{\bibinfo}[2]{#2}
\providecommand{\BIBentrySTDinterwordspacing}{\spaceskip=0pt\relax}
\providecommand{\BIBentryALTinterwordstretchfactor}{4}
\providecommand{\BIBentryALTinterwordspacing}{\spaceskip=\fontdimen2\font plus
\BIBentryALTinterwordstretchfactor\fontdimen3\font minus
  \fontdimen4\font\relax}
\providecommand{\BIBforeignlanguage}[2]{{%
\expandafter\ifx\csname l@#1\endcsname\relax
\typeout{** WARNING: IEEEtran.bst: No hyphenation pattern has been}%
\typeout{** loaded for the language `#1'. Using the pattern for}%
\typeout{** the default language instead.}%
\else
\language=\csname l@#1\endcsname
\fi
#2}}
\providecommand{\BIBdecl}{\relax}
\BIBdecl

\bibitem{chambers2009unsupervised}
N.~Chambers and D.~Jurafsky, ``Unsupervised learning of narrative schemas and
  their participants,'' in \emph{Proceedings of the Joint Conference of the
  47th Annual Meeting of the ACL and the 4th International Joint Conference on
  Natural Language Processing of the AFNLP}, 2009, pp. 602--610.

\bibitem{barzilay2008modeling}
R.~Barzilay and M.~Lapata, ``Modeling local coherence: An entity-based
  approach,'' \emph{Computational Linguistics}, vol.~34, no.~1, pp. 1--34,
  2008.

\bibitem{lapata2003probabilistic}
M.~Lapata, ``Probabilistic text structuring: Experiments with sentence
  ordering,'' in \emph{Proceedings of the 41st Annual Meeting of the
  Association for Computational Linguistics}, 2003, pp. 545--552.

\bibitem{roemmele2018neural}
M.~Roemmele, ``Neural networks for narrative continuation,'' Ph.D.
  dissertation, University of Southern California, 2018.

\bibitem{barzilay2002inferring}
R.~Barzilay and N.~Elhadad, ``Inferring strategies for sentence ordering in
  multidocument news summarization,'' \emph{Journal of Artificial Intelligence
  Research}, vol.~17, pp. 35--55, 2002.

\bibitem{roemmele2015creative}
M.~Roemmele and A.~S. Gordon, ``Creative help: a story writing assistant,'' in
  \emph{International Conference on Interactive Digital Storytelling}.\hskip
  1em plus 0.5em minus 0.4em\relax Springer, 2015, pp. 81--92.

\bibitem{yu2018qanet}
A.~W. Yu, D.~Dohan, M.-T. Luong, R.~Zhao, K.~Chen, M.~Norouzi, and Q.~V. Le,
  ``Qanet: Combining local convolution with global self-attention for reading
  comprehension,'' \emph{arXiv preprint arXiv:1804.09541}, 2018.

\bibitem{wang2020sbert}
B.~Wang and C.-C.~J. Kuo, ``Sbert-wk: A sentence embedding method by dissecting
  bert-based word models,'' \emph{arXiv preprint arXiv:2002.06652}, 2020.

\bibitem{sravanthi2017semantic}
P.~Sravanthi and B.~Srinivasu, ``Semantic similarity between sentences,''
  \emph{International Research Journal of Engineering and Technology (IRJET)},
  vol.~4, no.~1, pp. 156--161, 2017.

\bibitem{mostafazadeh2016corpus}
N.~Mostafazadeh, N.~Chambers, X.~He, D.~Parikh, D.~Batra, L.~Vanderwende,
  P.~Kohli, and J.~Allen, ``A corpus and cloze evaluation for deeper
  understanding of commonsense stories,'' in \emph{Proceedings of the 2016
  Conference of the North American Chapter of the Association for Computational
  Linguistics: Human Language Technologies}, 2016, pp. 839--849.

\bibitem{barzilay2004catching}
R.~Barzilay and L.~Lee, ``Catching the drift: Probabilistic content models,
  with applications to generation and summarization,'' \emph{arXiv preprint
  cs/0405039}, 2004.

\bibitem{gong2016end}
J.~Gong, X.~Chen, X.~Qiu, and X.~Huang, ``End-to-end neural sentence ordering
  using pointer network,'' \emph{arXiv preprint arXiv:1611.04953}, 2016.

\bibitem{agrawal2016sort}
H.~Agrawal, A.~Chandrasekaran, D.~Batra, D.~Parikh, and M.~Bansal, ``Sort
  story: Sorting jumbled images and captions into stories,'' \emph{arXiv
  preprint arXiv:1606.07493}, 2016.

\bibitem{li2016neural}
J.~Li and D.~Jurafsky, ``Neural net models for open-domain discourse
  coherence,'' \emph{arXiv preprint arXiv:1606.01545}, 2016.

\bibitem{sutskever2014sequence}
I.~Sutskever, O.~Vinyals, and Q.~V. Le, ``Sequence to sequence learning with
  neural networks,'' in \emph{Advances in neural information processing
  systems}, 2014, pp. 3104--3112.

\bibitem{chen2016neural}
X.~Chen, X.~Qiu, and X.~Huang, ``Neural sentence ordering,'' \emph{arXiv
  preprint arXiv:1607.06952}, 2016.

\bibitem{prabhumoye2020topological}
S.~Prabhumoye, R.~Salakhutdinov, and A.~W. Black, ``Topological sort for
  sentence ordering,'' \emph{arXiv preprint arXiv:2005.00432}, 2020.

\bibitem{kumar2020deep}
P.~Kumar, D.~Brahma, H.~Karnick, and P.~Rai, ``Deep attentive ranking networks
  for learning to order sentences.'' in \emph{AAAI}, 2020, pp. 8115--8122.

\bibitem{yin2020enhancing}
Y.~Yin, F.~Meng, J.~Su, Y.~Ge, L.~Song, J.~Zhou, and J.~Luo, ``Enhancing
  pointer network for sentence ordering with pairwise ordering predictions.''
  in \emph{AAAI}, 2020, pp. 9482--9489.

\bibitem{yin2019graph}
Y.~Yin, L.~Song, J.~Su, J.~Zeng, C.~Zhou, and J.~Luo, ``Graph-based neural
  sentence ordering,'' \emph{arXiv preprint arXiv:1912.07225}, 2019.

\bibitem{wang2019hierarchical}
T.~Wang and X.~Wan, ``Hierarchical attention networks for sentence ordering,''
  in \emph{Proceedings of the AAAI Conference on Artificial Intelligence},
  vol.~33, 2019, pp. 7184--7191.

\bibitem{oh2019topic}
B.~Oh, S.~Seo, C.~Shin, E.~Jo, and K.-H. Lee, ``Topic-guided coherence modeling
  for sentence ordering by preserving global and local information,'' in
  \emph{Proceedings of the 2019 Conference on Empirical Methods in Natural
  Language Processing and the 9th International Joint Conference on Natural
  Language Processing (EMNLP-IJCNLP)}, 2019, pp. 2273--2283.

\bibitem{bohn2018learning}
T.~Bohn, Y.~Hu, J.~Zhang, and C.~X. Ling, ``Learning sentence embeddings for
  coherence modelling and beyond,'' \emph{arXiv preprint arXiv:1804.08053},
  2018.

\bibitem{mostafazadeh2017lsdsem}
N.~Mostafazadeh, M.~Roth, A.~Louis, N.~Chambers, and J.~Allen, ``Lsdsem 2017
  shared task: The story cloze test,'' in \emph{Proceedings of the 2nd Workshop
  on Linking Models of Lexical, Sentential and Discourse-level Semantics},
  2017, pp. 46--51.

\bibitem{clark2016improving}
K.~Clark and C.~D. Manning, ``Improving coreference resolution by learning
  entity-level distributed representations,'' \emph{arXiv preprint
  arXiv:1606.01323}, 2016.

\bibitem{zhelezniak2019correlation}
V.~Zhelezniak, A.~Savkov, A.~Shen, and N.~Y. Hammerla, ``Correlation
  coefficients and semantic textual similarity,'' \emph{arXiv preprint
  arXiv:1905.07790}, 2019.

\bibitem{cer2018universal}
D.~Cer, Y.~Yang, S.-y. Kong, N.~Hua, N.~Limtiaco, R.~S. John, N.~Constant,
  M.~Guajardo-Cespedes, S.~Yuan, C.~Tar \emph{et~al.}, ``Universal sentence
  encoder,'' \emph{arXiv preprint arXiv:1803.11175}, 2018.

\bibitem{devlin2018bert}
J.~Devlin, M.-W. Chang, K.~Lee, and K.~Toutanova, ``Bert: Pre-training of deep
  bidirectional transformers for language understanding,'' \emph{arXiv preprint
  arXiv:1810.04805}, 2018.

\bibitem{jiang2020combining}
Z.~Jiang, M.~Srivastava, S.~Krishna, D.~Akodes, and R.~Schwartz, ``Combining
  word embeddings and n-grams for unsupervised document summarization,''
  \emph{arXiv preprint arXiv:2004.14119}, 2020.

\bibitem{mikolov2013efficient}
T.~Mikolov, K.~Chen, G.~Corrado, and J.~Dean, ``Efficient estimation of word
  representations in vector space,'' \emph{arXiv preprint arXiv:1301.3781},
  2013.

\bibitem{kiros2015skip}
R.~Kiros, Y.~Zhu, R.~R. Salakhutdinov, R.~Zemel, R.~Urtasun, A.~Torralba, and
  S.~Fidler, ``Skip-thought vectors,'' in \emph{Advances in neural information
  processing systems}, 2015, pp. 3294--3302.

\bibitem{lin2004automatic}
C.-Y. Lin and F.~J. Och, ``Automatic evaluation of machine translation quality
  using longest common subsequence and skip-bigram statistics,'' in
  \emph{Proceedings of the 42nd Annual Meeting of the Association for
  Computational Linguistics (ACL-04)}, 2004, pp. 605--612.

\bibitem{logeswaran2016sentence}
L.~Logeswaran, H.~Lee, and D.~Radev, ``Sentence ordering and coherence modeling
  using recurrent neural networks,'' \emph{arXiv preprint arXiv:1611.02654},
  2016.

\bibitem{cui2018deep}
B.~Cui, Y.~Li, M.~Chen, and Z.~Zhang, ``Deep attentive sentence ordering
  network,'' in \emph{Proceedings of the 2018 Conference on Empirical Methods
  in Natural Language Processing}, 2018, pp. 4340--4349.

\bibitem{lapata2006automatic}
M.~Lapata, ``Automatic evaluation of information ordering: Kendall's tau,''
  \emph{Computational Linguistics}, vol.~32, no.~4, pp. 471--484, 2006.

\bibitem{cormen2009introduction}
T.~H. Cormen, C.~E. Leiserson, R.~L. Rivest, and C.~Stein, \emph{Introduction
  to algorithms}.\hskip 1em plus 0.5em minus 0.4em\relax MIT press, 2009.

\end{thebibliography}

\end{document}